\documentclass[12pt]{article}

\usepackage[colorlinks,allcolors=blue,bookmarks=false,hypertexnames=true]{hyperref}
\usepackage[small,bf]{caption}
\usepackage{fullpage}
\usepackage{graphicx}
\usepackage{amsmath}
\usepackage{url}
\usepackage{amssymb}
\usepackage{booktabs}

\newcommand{\ones}{\mathbf 1}
\newcommand{\reals}{{\mbox{\bf R}}}
\newcommand{\complexes}{{\mbox{\bf C}}}

\newcommand{\symm}{{\mbox{\bf S}}}  

\newcommand{\Tr}{\mathop{\bf Tr}}
\newcommand{\diag}{\mathop{\bf diag}}

\newcommand{\prox}{\mathbf{prox}}

\newcommand{\argmin}{\mathop{\rm argmin}}

\newcommand{\ie}{{\it i.e.}}
\newcommand{\etc}{{\it etc.}}

\newcommand{\BEAS}{\begin{eqnarray*}}
\newcommand{\EEAS}{\end{eqnarray*}}
\newcommand{\BEA}{\begin{eqnarray}}
\newcommand{\EEA}{\end{eqnarray}}
\newcommand{\BEQ}{\begin{equation}}
\newcommand{\EEQ}{\end{equation}}
\newcommand{\BIT}{\begin{itemize}}
\newcommand{\EIT}{\end{itemize}}

\renewcommand{\hat}{\widehat}
\renewcommand{\tilde}{\widetilde}

\newcounter{algorithmctr}[section]
\renewcommand{\thealgorithmctr}{\thesection.\arabic{algorithmctr}}
\newenvironment{algdesc}%
   {\refstepcounter{algorithmctr}\begin{list}{}{%
       \setlength{\rightmargin}{0\linewidth}%
       \setlength{\leftmargin}{.05\linewidth}}%
       \rmfamily\small
       \item[]{\setlength{\parskip}{0ex}\hrulefill\par%
        \nopagebreak{\bfseries\textsf{Algorithm \thealgorithmctr~}}}}%
   {{\setlength{\parskip}{-1ex}\nopagebreak\par\hrulefill} \end{list}}

\title{Fitting Laplacian Regularized\\ Stratified Gaussian Models}
\author{Jonathan Tuck \and Stephen Boyd}

\begin{document}
\maketitle

\begin{abstract}
We consider the problem of jointly estimating multiple related 
zero-mean Gaussian distributions from data.
We propose to jointly estimate these covariance matrices using 
Laplacian regularized stratified model fitting, which includes
loss and regularization terms for each covariance matrix,
and also a term that encourages the different 
covariances matrices to be close.
This method `borrows strength' from the neighboring 
covariances, to improve its estimate.
With well chosen hyper-parameters, such models can perform
very well, especially in the low data regime.
We propose a distributed method that scales to large problems, and 
illustrate the efficacy of the method with examples
in finance, radar signal processing, and weather forecasting.
\end{abstract}

\section{Introduction}\label{s:intro}

We observe data records of the form $(z, y)$, where $y \in \reals^n$ 
and $z \in \{1, \ldots, K\}$.
We model $y$ as samples from a zero-mean Gaussian
distribution, conditioned on $z$, \ie,
\[
y \mid z \sim \mathcal N(0, \Sigma_z),
\]
with $\Sigma_z \in \symm_{++}^n$ (the set of symmetric positive definite
$n \times n$ matrices), $z=1, \ldots, K$.
Our goal is to estimate the model parameters 
$\Sigma = (\Sigma_1, \ldots, \Sigma_K)\in (\symm_{++}^n)^K$ from the data.
We refer to this as a stratified Gaussian model, since
we have a different Gaussian model for $y$ for each value of the 
stratification feature $z$.
Estimating a set of covariance matrices is referred to as 
joint covariance estimation.

The negative log-likelihood of $\Sigma$, on an observed data set
$(z_i,y_i)$, $i=1, \ldots, m$, is given by
\BEAS
\lefteqn{\sum_{i=1}^m \left( 
 (1/2) y_i^T\Sigma_{z_i}^{-1}y_i - (1/2)\log\det(\Sigma_{z_i}^{-1}) - (n/2)\log(2\pi) 
 \right)}\\
&=&\sum_{k=1}^K \left(
 (n_k/2)\Tr(S_k \Sigma_{k}^{-1}) - (n_k/2)\log\det(\Sigma_k^{-1}) - (n_k n/2)\log(2\pi) 
 \right),
\EEAS
where $n_k$ is the number of data samples with $z=k$
and $S_k = \frac{1}{n_k}\sum_{i : z_i = k} y_i y_i^T$ is the empirical covariance matrix
of $y$ for which $z=k$, with $S_k=0$ when $n_k=0$.

This function is general not convex in $\Sigma$, 
but it is convex in the \emph{natural parameter} 
\[
\theta = (\theta_1, \ldots, \theta_K) \in (\symm_{++}^n)^K,
\]
where $\theta_k = \Sigma_k^{-1}$, $k=1\ldots, K$.
We will focus on estimating $\theta$ rather than $\Sigma$.
In terms of $\theta$, and dropping a constant and a factor of two,
the negative log-likelihood is
\[
\ell(\theta) 
= \sum_{k=1}^K \ell_k(\theta_k),
\]
where
\[
\ell_k (\theta_k) = n_k\left( \Tr(S_k \theta_k) - \log\det(\theta_k) \right).
\]
We refer to $\ell(\theta)$ as the loss, and $\ell_k(\theta_k)$
as the local loss, associated with $z=k$.
For the special case where $n_k=0$, we define $\ell_k(\theta_k)$
to be zero if $\theta_k \succ 0$, and $\infty$ otherwise.
We refer to $\ell(\theta)/m$ as the average loss.

To estimate $\theta$, we add two types of regularization to the loss, and 
minimize the sum.
We choose $\theta$ as a solution of
\BEQ
\begin{array}{ll}
\mbox{minimize} &
\sum_{k=1}^K \left( {\ell_k}(\theta_k) + r(\theta_k) \right)
+ \mathcal L(\theta),
\end{array}
\label{eq:strat-obj}
\EEQ
where $\theta$ is the optimization variable,
$r : \symm^n \to \reals$ is a local regularization function,
and $\mathcal L : (\symm^n)^K \to \reals$ is Laplacian regularization,
defined below.
We refer to our estimated $\theta$ as a Laplacian 
regularized stratified Gaussian model.

\paragraph{Local regularization.}
Common types of local regularization include
trace regularization, $r(\theta_k)= \gamma \Tr \theta_k$,
and Frobenius regularization,
$r(\theta_k)= \gamma \| \theta_k \|_F^2$,
where $\gamma>0$ is a hyper-parameter.
\nocite{fazel2002nuclearnorm, recht2010nuclearnorm}
Two more recently introduced local regularization terms are
$\gamma \|\theta_k\|_1$
and $\gamma \|\theta_k\|_{\mathrm{od},1} 
= \gamma \sum_{i \neq j} |(\theta_k)_{ij}|$, which encourage 
sparsity of $\theta$ and of the off-diagonal elements of 
$\theta$, respectively \cite{friedman2008sparse}.
(A zero entry in $\theta_k$ means that the associated components of $y$ are 
conditionally indedendent, given the others, when $z=k$
\cite{danaher2014jointgraphicallasso}.)

\paragraph{Laplacian regularization.}
Let $W \in \symm^K$ be a symmetric matrix with zero diagonal entries 
and nonnegative off-diagonal entries.
The associated Laplacian regularization is the function
$\mathcal L: (\symm^n)^K \to \reals$ given by
\[
\mathcal L(\theta)
= \frac{1}{2}\sum_{i,j=1}^K W_{ij} \|\theta_i - \theta_j\|_F^2.
\]
Evidently $\mathcal L$ is separable across the entries of 
its arguments; it can be expressed as
\[
\mathcal L (\theta) = \frac{1}{2}\sum_{u,v=1}^n  \left( 
\sum_{i,j=1}^K W_{ij} ((\theta_i)_{uv}-(\theta_j)_{uv})^2 \right).
\]
Laplacian regularization encourages the estimated values of
$\theta_i$ and $\theta_j$ to be close when $W_{ij}>0$.
Roughly speaking, we can interpret $W_{ij}$ as prior knowledge
about how close the data generation processes for $y$ are,
for $z=i$ and $z=j$.

We can associate the Laplacian regularization with a graph with $K$ 
vertices, which has an edge $(i,j)$ for each positive $W_{ij}$, with 
weight $W_{ij}$.
We refer to this graph as the regularization graph.
We assume that the regularization graph is connected.
We can express Laplacian regularization in terms of a (weighted) 
Laplacian matrix $L$, given by
\[
L_{ij}  = \left\{ \begin{array}{ll}
-W_{ij} & i \neq j \\
\sum_{k=1}^K W_{ik} & i=j
\end{array} \right.
\]
for $i,j=1,\ldots, K$.
The Laplacian regularization can be expressed in terms of $L$ as 
\[
{\mathcal L}(\theta) = (1/2) \Tr(\theta^T (I \otimes L) \theta),
\] 
where $\otimes$ denotes the Kronecker product.

\paragraph{Assumptions.}
We note that~(\ref{eq:strat-obj}) need not have a unique solution, in pathological
cases.  
As a simple example, consider the case with $r=0$ and $W=0$, \ie, no
local regularization and no Laplacian regularization, which corresponds to
independently creating a model for each value of $z$.  
If all $S_k$ are positive definite, the solution is unique, with $\theta_k = S_k^{-1}$.  
If any $S_k$ is not positive definite, the problem does not have a unique solution.
The presence of either local or Laplacian regularization (with the associated graph
being connected) can ensure that the problem has a unique solution.  
For example, with trace regularization (and $\gamma>0$), it is readily shown that 
the problem~(\ref{eq:strat-obj}) has a unique solution.  
Another elementary condition that guarantees a unique solution is that the associated 
graph is connected, and $S_k$ do not have a common nullspace.

We will henceforth assume that the problem~\eqref{eq:strat-obj} has a unique solution.
This implies that the objective in~\eqref{eq:strat-obj}
is closed, proper, and convex.
The problem \eqref{eq:strat-obj} is a convex optimization problem which can be solved 
globally in an efficient manner~\cite{vandenberghe1996semidefinite,boyd2004convex}.

\paragraph{Contributions.}
Joint covariance estimation and 
Laplacian regularized stratified model fitting are not new ideas;
in this paper we simply bring them together.
Laplacian regularization has been shown to work well in conjunction
with stratified models, allowing one with very little data to create sensible 
models for each value of some stratification 
parameter~\cite{tuck2019stratmodels,tuck2020eigenstrat}.
To our knowledge, this is the first paper that has explicitly framed joint 
covariance estimation as a stratified model fitting problem.
We develop and implement a large-scale distributed method for 
Laplacian regularized joint covariance estimation via the alternating 
direction method of multipliers (ADMM), 
which scales to large-scale data sets 
\cite{boyd2011distributed, wahlberg2012admmtvest}.

\paragraph{Outline.}
In \S\ref{s:intro}, we introduce Laplacian regularized 
stratified Gaussian models and review work related to fitting 
Laplacian regularized stratified Gaussian models.
In \S\ref{s:soln_method}, we develop and analyze a distributed
solution method to fit Laplacian regularized stratified Gaussian
models, based on ADMM.
Lastly, in \S\ref{s:examples}, we illustrate the efficacy of this model
fitting technique and of this method with three examples,
in finance, radar signal processing, and weather forecasting.

\subsection{Related work}\label{ss:related_work}

\paragraph{Stratified model fitting.}
Stratified model fitting, \ie, separately fitting a different model for 
each value of some parameter, is an idea widely used across disciplines.
For example, in medicine, patients are often divided into subgroups based 
on age and sex, and one fits a separate model for the data from each 
subgroup~\cite{kernan1999stratified,tuck2019stratmodels}.
Stratification can be useful for dealing with categorical feature values, 
interpreting the nature of the data, and can play a large role in experiment
design.
As mentioned previously, the joint covariance estimation problem 
can naturally be framed as a stratified model fitting problem.

\paragraph{Covariance matrix estimation.}
Covariance estimation applications span disciplines
such as radar signal processing \cite{salari2019jointcovest}, 
statistical learning \cite{banerjee2008sparsemle},
finance \cite{almgren2000portfolio,skaf2009mpo}, 
and medicine \cite{levitan1987mapcov}.
Many techniques exist for the estimation of a single covariance 
matrix when the 
covariance matrix's structure is known \emph{a priori} 
\cite{fan2016covest}.
When the covariance matrix is sparse, thresholding the 
elements of the sample covariance matrix has been shown to be an 
effective method of covariance matrix estimation 
\cite{bickel2008covest}.
\cite{steiner2000covmtx} propose a maximum likelihood solution
for a covariance matrix that is the sum of a Hermitian positive 
semidefinite matrix and a multiple of the identity.
Maximimum likelihood-style approaches also exist for when the 
covariance matrix is assumed to be Hermitian, Toeplitz, or both 
\cite{burg1982covmtx,miller1987covmtx,li1999covmtx}.
\cite{cao2009covmtx} propose using various shrinkage estimators when 
the data is high dimensional.
(Shrinkage parameters are typically chosen by an out-of-sample 
validation technique \cite{hoffbeck1996covmtx}.)

\paragraph{Joint covariance estimation.}
Jointly estimating statistical model parameters has been the 
subject of significant research spanning different disciplines.
The joint graphical lasso \cite{danaher2014jointgraphicallasso} 
is a stratified model that encourages closeness of parameters by 
their difference as measured by fused lasso and group lasso penalties.
(Laplacian regularization penalizes their difference by the 
$\ell_2$-norm squared.)
The joint graphical lasso penalties in effect result in groups 
of models with the same parameters, and those parameters being sparse.
(In contrast, Laplacian regularization leads to parameter values that 
vary smoothly with nearby models.
It has been observed that in most practical settings, Laplacian 
regularization is sufficient for accurate estimation \cite{tuck2019stratmodels}.)
Similar to the graphical lasso, methods such as the time-varying graphical lasso
\cite{hallac2017network} and the network lasso \cite{hallac2015network}
have been recently developed to infer model parameters in graphical networks
assuming some graphical relationship (in the former, the relationship is in time;
in the latter, the relationship is arbitrary).

Another closely related work to this paper is \cite{tuck2018distributed}, 
which introduces the use of Laplacian regularization in joint estimation 
of covariance matrices in a zero-mean multivariate Gaussian model.
In this paper, Laplacian regularization is used assuming a grid structure, 
and the problem is solved using the majorization-minimization algorithmic 
framework~\cite{SBP:17}.
In contrast, this paper assumes a much more complex and sophisticated 
structure of the system, and uses ADMM to solve the problem much more efficiently.

\paragraph{Connection to probabilistic graphical models.}
There is a significant connection of this work to probabilistic 
graphical models \cite{koller2009probgraphmodels}.
In this connection, a stratified model for joint model parameter 
estimation can be seen as an undirected graphical model, where 
the vertices follow different distributions, and the edges encourage 
corresponding vertices' distributions to be alike.
In fact, very similar problems in atmospheric science, medicine, 
and statistics have been studied under this 
context
\cite{guo2011jointgraphical,danaher2014jointgraphicallasso,
zhu2014jointgraphical,ma2016jointgraphical}.

\section{Distributed solution method}\label{s:soln_method}
There are many methods that can be used to solve minimize \eqref{eq:strat-obj}; 
for example, ADMM~\cite{boyd2011distributed} has been successfully used in the 
past as a large-scale, distributed method for stratified model fitting with
Laplacian regularization \cite{tuck2019stratmodels}, which we will adapt for 
use in this paper.
This method expresses minimizing~\eqref{eq:strat-obj} in the equivalent form
\BEQ
\begin{array}{ll}
\label{eq:ADMM_jointcovest}
{{\mbox{minimize}}} & \overset{K}{\underset{k=1}\sum} 
(({\ell_k}(\theta_k) + r(\tilde{\theta}_k))
+ \mathcal L(\hat{\theta})\\
{\mbox{subject to}} 
& \theta-\hat{\theta}=0, 
\quad \tilde{\theta} - \hat{\theta}=0,
\end{array}
\EEQ
now with variables $\theta \in (\symm_{++}^n)^K$, 
$\tilde{\theta} \in (\symm_{++}^n)^K$, 
and $\hat{\theta} \in (\symm_{++}^n)^K$.
Problem~\eqref{eq:ADMM_jointcovest} is in ADMM standard form, splitting
on $(\theta,\tilde\theta)$ and $\hat\theta$.
The ADMM algorithm for this problem, outlined in full in Algorithm~\ref{a-dist_strat}, 
can be summarized by four steps: computing the (scaled) proximal operators of 
$\ell_1, \ldots, \ell_K$, $r$, and $\mathcal L$,
followed by updates on dual variables associated with the two
equality constraints, $U \in (\reals^{n \times n})^K$ and 
${\tilde U} \in (\reals^{n \times n})^K$.
Recall that the proximal operator of $f: \reals^{n \times n} \to \reals$ 
with penalty parameter $\omega$ is
\[
\prox_{\omega f}(V) = \underset{\theta}\argmin
\left(\omega f(\theta) + (1/2)\|\theta-V\|_F^2\right).
\]

\begin{algdesc}
\label{a-dist_strat}
\emph{Distributed method for Laplacian regularized joint covariance estimation.}
\begin{tabbing}
 {\bf given} Loss functions $\ell_1,\ldots,\ell_K$, 
 local regularization function $r$,
 graph Laplacian matrix $L$,\\
 and penalty parameter $\omega>0$.\\
 \emph{Initialize}.
 $\theta^0={\tilde\theta}^0={\hat{\theta}}^0=U^0={\tilde U}^0=0$.\\
 {\bf repeat} \\
  \qquad \=\ 1.\
 \emph{Evaluate the proximal operator of ${\ell_k}$.}
 $\theta^{t+1}_k = \prox_{\omega \ell_k}({\hat{\theta}}_k^t - U_k^t), 
 \quad k = 1,\ldots,K$\\
  \qquad \=\ 2.\
 \emph{Evaluate the proximal operator of $r$.}
 ${\tilde\theta}^{t+1}_k 
 = \prox_{\omega r}({\hat{\theta}}_k^t - {\tilde U}_k^t), 
 \quad k = 1,\ldots,K$\\
 \qquad \=\ 3.\
 \emph{Evaluate the proximal operator of $\mathcal L$.}
 ${\hat \theta}^{t+1} = \prox_{\omega {\mathcal L} / 2}((1/2)(\theta^{t+1}+U^t+{\tilde\theta}^{t+1}+{\tilde U}^{t}))$\\
 \qquad \=\ 4.\
 \emph{Update the dual variables.}
 ${{U}}^{t+1} = {{U}}^t + {\theta}^{t+1} - {\hat{\theta}}^{t+1}; 
 \quad {\tilde{U}}^{t+1}
 = {\tilde{U}}^t + \tilde{\theta}^{t+1} - {\hat{\theta}}^{t+1}$\\
 {\bf until convergence}
\end{tabbing}
\end{algdesc}

To see how we could use this for fitting Laplacian regularized 
stratified models for the joint covariance estimation problem,
we outline efficient methods for evaluating the proximal operators of $\ell_k$, 
of a variety of relevant local regularizers $r$, and of the Laplacian regularization.

\subsection{Evaluating the proximal operator of $\ell_k$}
Evaluating the proximal operator of $\ell_k$ (for $n_k > 0$) can be done efficiently and 
in closed-form \cite{WT:09,boyd2011distributed,danaher2014jointgraphicallasso,
tuck2019stratmodels}.
We have that the proximal operator is
\[
\prox_{\omega \ell_k}(V) = Q X Q^T,
\]
where $X \in \reals^{K \times K}$ is a diagonal matrix with entries
\[
X_{ii} = \frac{\omega n_k d_i + \sqrt{(\omega n_k d_i)^2 + 4\omega n_k} }{2},
\quad i = 1, \ldots, K,
\]
and $d$ and $Q$ are computed as the eigen-decomposition of $(1/\omega n_k)V - S_k$,
\ie,
\[
\frac{1}{\omega n_k}V - S_k = Q \diag(d) Q^T.
\]
The dominant cost in computing the proximal operator of $\ell_k$ is in computing the
eigen-decomposition, which can be computed with order $n^3$ flops.

\subsection{Evaluating the proximal operator of $r$}
The proximal operator of $r$ often has a closed-form expression
that can be computed in parallel.
For example, if $r=\gamma\Tr(\theta)$, then $\prox_{\omega r}(V) = V - \omega \gamma I$.
If $r(\theta) = (\gamma/2)\|\theta\|_F^2$ then $\prox_{\omega r}(V) = (1/(1+\omega\gamma))V$, 
and if $r = \gamma\|\theta\|_1$, then $\prox_{\omega r}(V) = \max(V-\omega\gamma, 0) - \max(-V-\omega\gamma, 0)$,
where $\max$ is taken elementwise \cite{PB:14}.
If $r(\theta) = \gamma_1 \Tr(\theta) + \gamma_2 \|\theta\|_{\mathrm{od},1}$ 
where $\|\theta\|_{\mathrm{od},1}=\sum_{i \neq j} |\theta_{ij}|$ is the $\ell_1$-norm of the
off diagonal elements of $\theta$, then
\[
\prox_{\omega r}(V)_{ij} = 
\begin{cases}
V_{ij} - \omega\gamma_1, & i = j \\
\max(V_{ij}-\omega\gamma_2, 0) - \max(-V_{ij}-\omega\gamma_2, 0) & i \neq j
\end{cases}.
\]

\subsection{Evaluating the proximal operator of $\mathcal L$}
Evaluating the proximal operator of $\mathcal L$ is equivalent to
solving the $n(n+1)/2$ regularized Laplacian systems
\BEQ
\label{eq:laplacian_system}
\bigg(L+(2/\omega) I\bigg)
\begin{bmatrix}
(\hat\theta^{t+1}_1)_{ij} \\
(\hat\theta^{t+1}_2)_{ij} \\
\vdots \\
(\hat\theta^{t+1}_K)_{ij}
\end{bmatrix}
= (1/\omega)
\begin{bmatrix}
(\theta^{t+1}_1 + U^{t}_1 + {\tilde\theta}^{t+1}_1 + {\tilde U}^{t}_1)_{ij} \\
(\theta^{t+1}_2 + U^{t}_2 + {\tilde\theta}^{t+1}_2 + {\tilde U}^{t}_2)_{ij} \\
\vdots \\
(\theta^{t+1}_K + U^{t}_K + {\tilde\theta}^{t+1}_K + {\tilde U}^{t}_K)_{ij}
\end{bmatrix}
\EEQ
for $i=1,\ldots,n$ and $j = 1, \ldots, i$, and 
setting $({\hat \theta}_k^{t+1})_{ji} = ({\hat \theta}_k^{t+1})_{ij}$.
Solving these systems is quite efficient; many methods for solving 
Laplacian systems (and more generally, symmetric diagonally-dominant systems) 
can solve these systems in nearly-linear 
time~\cite{vishnoi2013lx, kelner2013sddsystems}.
We find that the conjugate gradient (CG) method with a diagonal 
pre-conditioner \cite{hestenes1952methods,takapoui2016preconditioning}
can efficiently and reliably solve these systems.
(We can also warm-start CG with $\hat\theta^t$.)

\paragraph{Stopping criterion.}
Under our assumptions on the objective, the iterates of ADMM converge 
to a global solution, and the primal and dual residuals
\[
r^{t+1} = (\theta^{t+1}-\hat\theta^{t+1},\tilde\theta^{t+1}-\hat\theta^{t+1}),\quad
s^{t+1} = -(1/\lambda)(\hat\theta^{t+1}-\hat\theta^{t},\hat\theta^{t+1}-\hat\theta^{t}),
\]
converge to zero \cite{boyd2011distributed}.
This suggests the stopping criterion
\[
\|r^{t+1}\|_F \leq \epsilon_\mathrm{pri}, \quad \|s^{t+1}\|_F \leq \epsilon_\mathrm{dual},
\]
for some primal tolerance $\epsilon_\mathrm{pri}$ 
and dual tolerance $\epsilon_\mathrm{dual}$.
Typically, these tolerances are selected as a combination of absolute
and relative tolerances; we use
\[
\epsilon_\mathrm{pri} 
= \sqrt{2Kn^2}\epsilon_\mathrm{abs} + \epsilon_\mathrm{rel} \max\{\|r^{t+1}\|_F,\|s^{t+1}\|_F\},
\quad \epsilon_\mathrm{dual} 
= \sqrt{2Kn^2}\epsilon_\mathrm{abs} + (\epsilon_\mathrm{rel}/\omega) \|(u^t,{\tilde u^t})\|_F,
\]
for some absolute tolerance $\epsilon_{\mathrm{abs}} > 0$
and relative tolerance $\epsilon_{\mathrm{rel}} > 0$.

\paragraph{Penalty parameter selection.}
In practice (\ie, in \S\ref{s:examples}), we find that the number of iterations to 
convergence does not change significantly with the choice of the penalty parameter 
$\omega$.
We found that fixing $\omega=0.1$ worked well across all of our experiments.

\section{Examples}\label{s:examples}
In this section we illustrate Laplacian regularized stratified 
model fitting for joint covariance estimation.
In each of the examples, we fit two models: 
a common model (a Gaussian model without stratification),
and a Laplacian regularized stratified Gaussian model.
For each model, we selected hyper-parameters that performed best 
under a validation technique.
We provide an open-source implementation of Algorithm~\ref{a-dist_strat},
along with the code used to create the examples, 
at \url{https://github.com/cvxgrp/strat_models}.
We train all of our models with an absolute tolerance $\epsilon_{\mathrm{abs}}=10^{-3}$
and a relative tolerance $\epsilon_{\mathrm{rel}}=10^{-3}$.
All computation was carried out on a 2014 MacBook Pro 
with four Intel Core i7 cores clocked at 3 GHz.

\subsection{Sector covariance estimation}\label{ss:finance}
Estimating the covariance matrix of a portfolio of time series is a central
task in quantitative finance, as it is a parameter to be estimated 
in the classical Markowitz portfolio 
optimization problem~\cite{markowitz1952portfolio,skaf2009mpo,boyd2017multiperiod}.
In addition, models for studying the dynamics of the variance of a time series 
(or multiple time series) data are common, such as with the GARCH 
family of models in statistics~\cite{engle1982garch}.
In this example, we consider the problem of modeling the 
covariance of daily sector returns, given market conditions observed 
the day prior.

\paragraph{Data records and dataset.}
We use daily returns from $n=9$ exchange-traded funds (ETFs) that cover the 
sectors of the stock market, measured daily, at close, from January 1, 2000
to January 1, 2018 (for a total of 4774 data points).
The ETFs used are XLB (materials), 
XLV (health care), XLP (consumer staples), 
XLY (consumer discretionary), XLE (energy), XLF (financials), 
XLI (industrials), XLK (technology), and XLU (utilities).
Each data record includes $y \in \reals^9$, the daily return of the sector ETFs.
The sector ETFs have individually been winsorized (clipped) 
at their 5th and 95th percentiles. 

Each data record also includes the market condition $z$,
which is derived from market indicators known on the day,
the five-day trailing averages of 
the market volume (as measured by the ETF SPY)
and volatility (as measured by the ticker VIX).
Each of these market indicators is binned into 2\% quantiles
(\ie, $0\%-2\%, 2\%-4\%, \ldots, 98\%-100\%$), 
making the number of stratification features 
$K = 50 \cdot 50 = 2500$.
We refer to $z$ as the market conditions.

We randomly partition the dataset into a training set consisting of 60\% of 
the data records, a validation set consisting of 20\% of the data records, 
and a held-out test set consisting of the remaining 20\% of the data records.
In the training set, there are an average of 1.2 data points per 
market condition, and the number of data points per market condition vary 
significantly.  
The most populated market condition contains 38 data points, 
and there are 1395 market conditions (more than half of the 2500 total)
for which there are zero data points.

\paragraph{Model.}
The stratified model in this case includes $K=2500$ different sector return 
(inverse) covariance matrices in $\symm_{++}^{9}$, indexed by the market conditions.
Our model has $Kn(n-1)/2=90000$ parameters.

\paragraph{Regularization.}
For local regularization, we use trace regularization with regularization 
weight $\gamma_{\mathrm{loc}}$, \ie, $r = \gamma_{\mathrm{loc}}\Tr(\cdot)$.

The regularization graph for the stratified model is the Cartesian 
product of two regularization graphs:
\begin{itemize}
\item \emph{Quantile of five-day trailing average volatility.} 
The regularization graph is a path graph with 50 vertices, 
with edge weights $\gamma_\mathrm{vix}$.
\item \emph{Quantile of five-day trailing average market volume.} 
The regularization graph is a path graph with 50 vertices, 
	with edge weights $\gamma_\mathrm{vol}$.
\end{itemize}
The corresponding Laplacian matrix has 12300 nonzero entries, with
hyper-parameters $\gamma_\mathrm{vix}$ and $\gamma_\mathrm{vol}$.
All together, our stratified Gaussian model has three hyper-parameters.

\paragraph{Results.}
We compared a stratified model to a common model.
The common model corresponds to solving one covariance estimation problem,
ignoring the market regime.

For the common model, we used $\gamma_{\mathrm{loc}}=5$.
For the stratified model, we used 
$\gamma_{\mathrm{loc}}=0.15$, $\gamma_{\mathrm{vix}}=1500$, 
and $\gamma_{\mathrm{vol}}=2500$.
These values were chosen based on a crude hyper-parameter search.
We compare the models' average loss over the held-out test 
set in table~\ref{tab:finance}.
We can see that the stratified model substantially outperforms the common model.
\begin{table}
\caption{Results for \S\ref{ss:finance}.}
 \vspace{.4em}
 \centering
 \begin{tabular}{lll}
   \toprule
   Model & Average test loss \\
   \midrule
   Common & $6.42 \times 10^{-3}$\\
   Stratified & $1.15 \times 10^{-3}$\\
   \bottomrule
 \end{tabular}
 \label{tab:finance}
\end{table}

\begin{figure}
\centering
 \includegraphics[width=.75\textwidth]{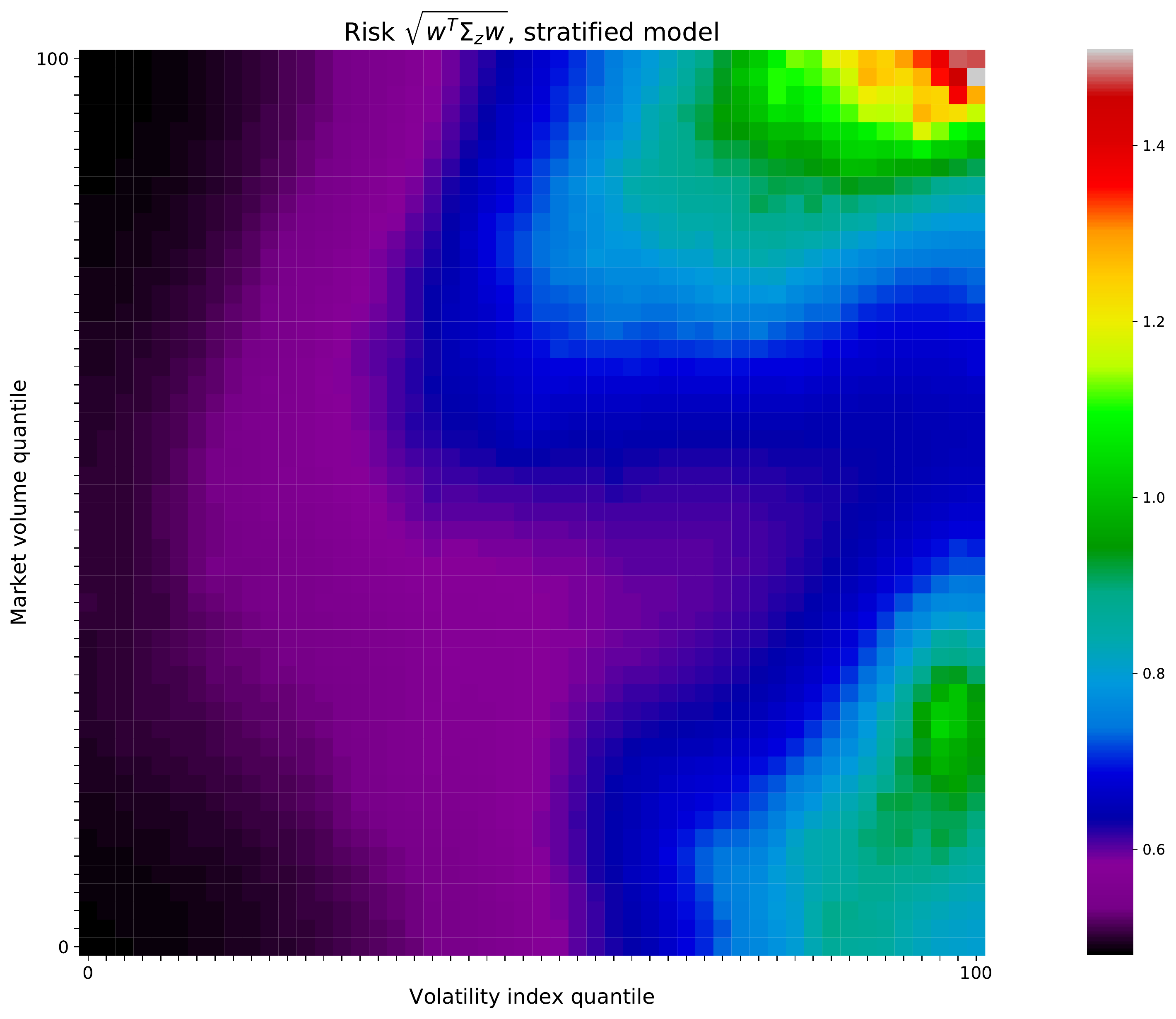}
   \caption{Heatmap of $\sqrt{w^T \Sigma_z w}$ with $w = (1/9)\ones$
   for the stratified model.}
\label{fig:risk_finance}
\end{figure}

To visualize how the covariance varies with market conditions,
we look at risk of a portfolio (\ie, the standard deviation of the return)
with uniform allocation across the sectors.
The risk is given by $\sqrt{w^T \Sigma w}$,
where $\Sigma$ is the covariance matrix and $w=(1/9)\ones$ is
the weight vector corresponding to a uniform allocation.
In figure~\ref{fig:risk_finance}, we plot the heatmap of the risk
of this portfolio as a function of the market regime $z$
for the stratified model.
The risk heatmap makes sense and varies smoothly across market conditions.
The estimate of the risk of the uniform portfolio 
for the common model covariance matrix is $0.859$.  
The risk in our stratified model varies by about a factor 
of two from this common estimate of risk.

\begin{figure}
  \centering
    \includegraphics[width=.75\textwidth]{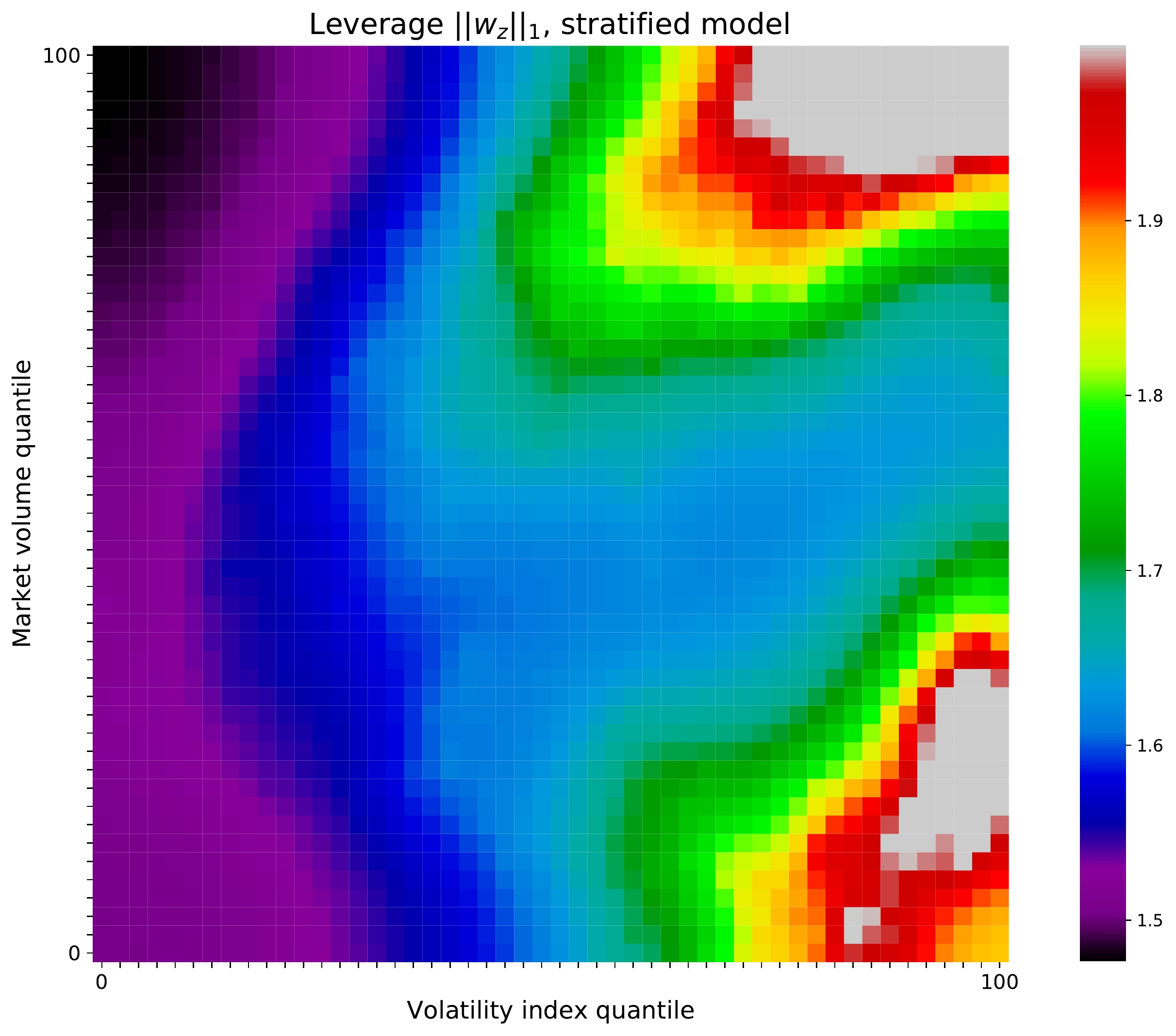}
      \caption{Heatmap of $\|w_z\|_1$, the stratified model portfolios, 
              indexed by market conditions.}
  \label{fig:leverage_finance}
  \end{figure}

\paragraph{Application.}
Here we demonstrate the use of our stratified risk model,
in a simple trading policy.
For each of the $K=2500$ market conditions, we compute the
portfolio $w_z \in \reals^9$ which is Markowitz optimal, 
\ie, the solution of
\BEA
\begin{array}{ll}
\mbox{maximize} & \mu^T w - \gamma w^T \Sigma_z w\\
\mbox{subject to} & \ones^T w = 1\\
                  & \| w \|_1 \leq 2,
\end{array}
\EEA
with optimization variable $w \in \reals^9$
($w_i < 0$ denotes a short position in asset $i$).
The objective is the risk adjusted return,
and $\gamma > 0$ is the \emph{risk-aversion parameter},
which we take as $\gamma=0.15$.
We take $\mu \in \reals^9$ to be the vector of median 
sector returns in the training set.
The last constraint limits the portfolio leverage, 
measured by $\|w\|_1$, to no more than 2.  
(This means that the 
total short positions in the portfolio cannot exceed $0.5$ times
the total portfolio value.)
We plot the leverage of the stratified model portfolios 
$w_z$, indexed by market conditions, in figure~\ref{fig:leverage_finance}.

\begin{figure}
  \centering
   \includegraphics[width=\textwidth]{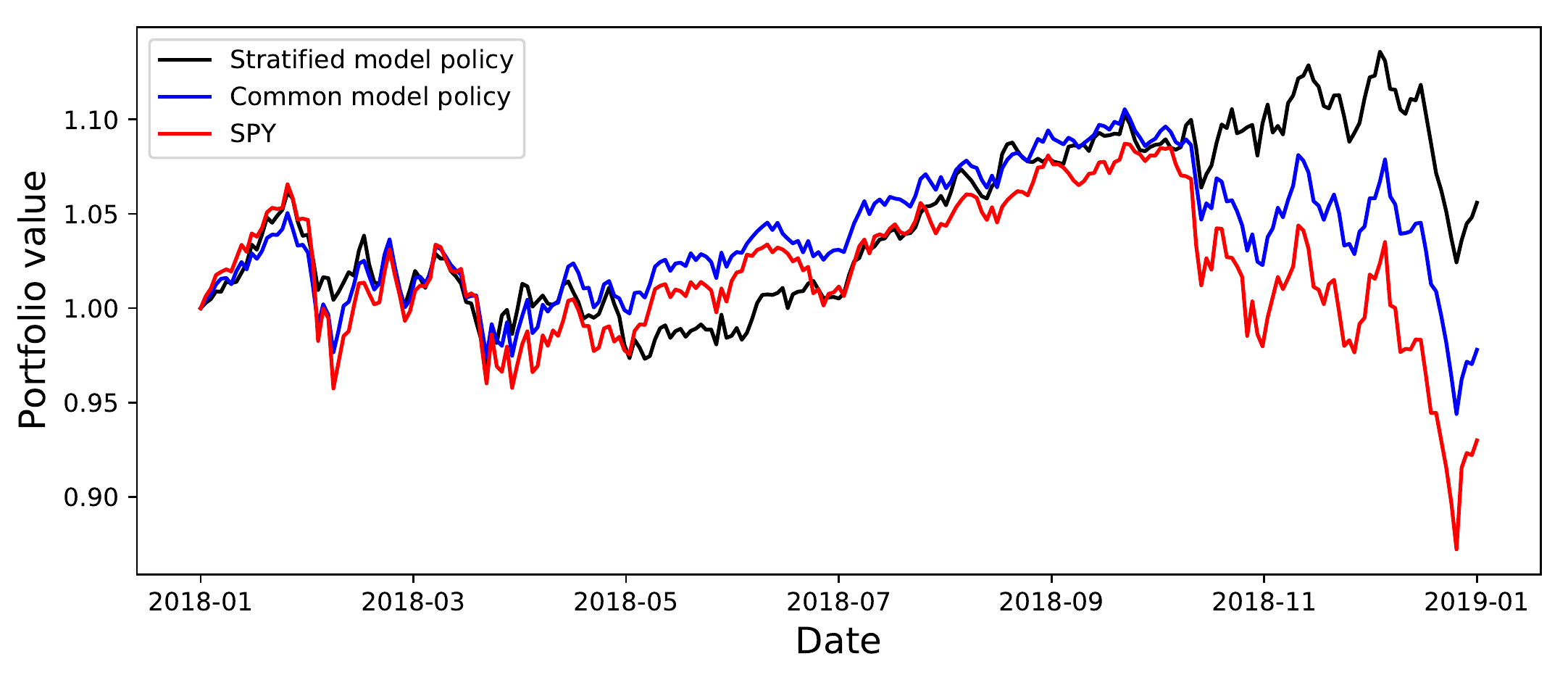}
     \caption{Cumulative value of three portfolios, 
starting from initial value $1$.}
  \label{fig:policy_finance}
  \end{figure}

\begin{figure}
  \centering
   \includegraphics[width=\textwidth]{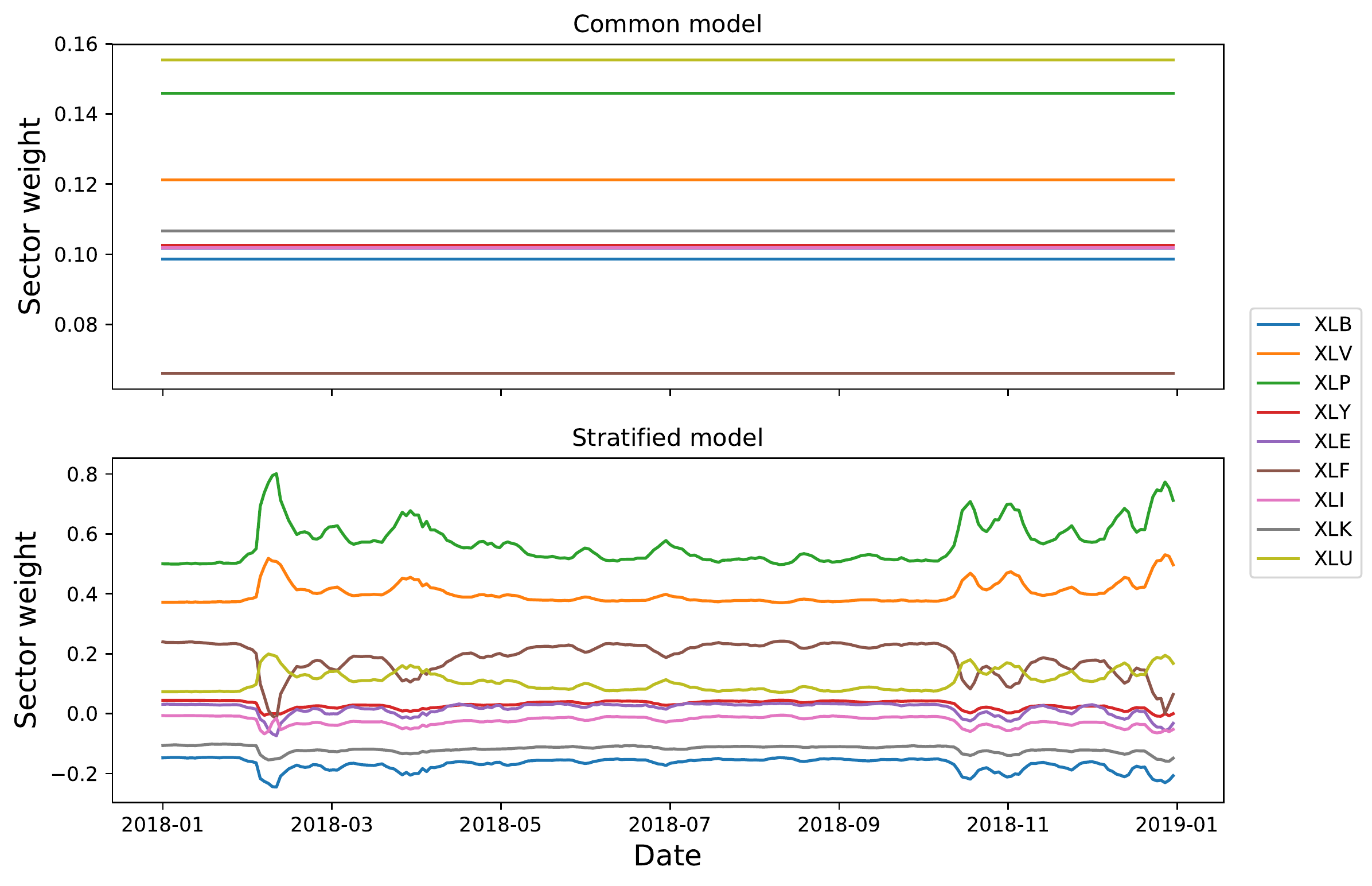}
     \caption{Weights for the stratified model policy and the common
policy over 2018.}
  \label{fig:weight_finance}
  \end{figure}

At the beginning of each day $t$, we use the previous day's market conditions 
$z_t$ to allocate our current total portfolio value according to the 
weights $w_{z_t}$. 
We run this policy using realized returns from January 1, 
2018 to January 1, 2019 
(which was held out from all other previous experiments).
In figure~\ref{fig:policy_finance}, we plot the cumulative value 
of three policies: 
Using the weights from the stratified model, using a constant 
weight from the common model, and simply buying and holding SPY.
In figure~\ref{fig:weight_finance}, we plot the sector weights of the 
stratified model policy and the weights of the common model policy.

Table~\ref{tab:ann_risk_ret_finance} gives the annualized realized 
risks and returns of the three policies.
The stratified model policy has both the lowest annualized risk 
and the greatest annualized return.  While the common model policy and 
buying and holding SPY realize losses over the year, our simple policy
has a positive realized return.
  \begin{table}
    \caption{Annualized realized risks and returns for 
    the three policies.}
      \vspace{.4em}
      \centering
      \begin{tabular}{lrr}
        \toprule
        Model & Annualized risk & Annualized return \\
        \midrule
        Stratified model & 0.112 & 0.061\\
        Common model & 0.124 & -0.015\\
        Buy/hold SPY & 0.170 & -0.059\\
        \bottomrule
      \end{tabular}
      \label{tab:ann_risk_ret_finance}
    \end{table}

\clearpage
\subsection{Space-time adaptive processing}\label{ss:radar}
In radar space time adaptive processing (STAP), a problem of widespread importance
is the detection problem: detect a target over a terrain in the presence of interference.
Interference typically comes in the form of clutter (unwanted terrain noise), 
jamming (noise emitted intentionally by an adversary), and white noise (typically caused 
by the circuitry/machinery of the radar receiver) \cite{melvin2004stap,wicks2006stap,
kang2015stap}.
(We refer to the sum of these three noises as interference.)
In practice, these covariance matrices for a given radar orientation 
(\ie, for a given range, azimuth, 
Doppler frequency, \etc) are unknown and must be estimated
\cite{wicks2006stap,kang2015stap}.
Our goal is to estimate the covariance matrix of the interference, given the 
radar orientation.

\paragraph{Data records.}
Our data records $(z,y)$ include ground interference measurements $y \in \reals^{30}$
(so $n=30$), which were synthetically generated (see below).
In addition, the stratification features $z$ describe the radar orientation.
A radar orientation corresponds to a tuple of the range $r$ (in km), 
azimuth angle $a$ (in degrees), and Doppler frequency $d$ (in Hz), which are binned.
For example, if $z = (r,a,d) = ([35, 37), [87, 89), [976, 980))$,
then the measurement was taken at a range between 35-57 km, 
an azimuth between 87-89 degrees, and a Doppler frequency between 
976-980 Hz.

There are 10 range bins, 10 azimuth bins, and 10 Doppler frequency bins,
and we allow $r \in [35, 50]$, $a \in [87, 267]$, 
and $d \in [-992, 992]$; these radar orientation values are realistic and 
were selected from the radar signal processing literature; see~\cite[Table 1]{radar_dataset}
and \cite[Table 3.1]{kang2015stap}.
The number of stratification features is $K=10 \cdot 10 \cdot 10 = 1000$.

We generated the data records $(z,y)$ as follows.
We generated three complex Hermitian matrices 
${\tilde \Sigma}_{\mathrm{range}} \in \complexes^{15}$, 
${\tilde \Sigma}_{\mathrm{azi}} \in \complexes^{15}$, 
and ${\tilde \Sigma}_{\mathrm{dopp}} \in \complexes^{15}$
randomly, where $\complexes$ is the set of complex numbers. 
For each $z=(r,a,d)$, we generate a covariance matrix according to
\[
  {\tilde \Sigma}_z 
= {\tilde \Sigma}_{(r,a,d)} 
= \left(\frac{4 \times 10^4}{r}\right)^2 {\tilde \Sigma}_{\mathrm{range}} 
+ \left(\cos\left(\frac{\pi a}{180}\right)+\sin\left(\frac{\pi a}{180}\right)\right) {\tilde \Sigma}_{\mathrm{azi}} 
+ \left(1+\frac{d}{1000}\right) {\tilde \Sigma}_{\mathrm{dopp}}.
\]
For each $z$, we then independently sample from a Gaussian distribution
with zero mean and covariance matrix ${\tilde \Sigma}_z$ to generate the corresponding 
data samples $\tilde y \in \reals^{15}$.
We then generate the real-valued data records $(z,y)$ from the complex-valued $(z, \tilde y)$
via $y = (\Re {\tilde y}, \Im {\tilde y})$, where $\Re$ and $\Im$ denote the real
and imaginary parts of $\tilde y$, respectively, and equivalently estimate (the inverses of)
\[
  \Sigma_z =
  \begin{bmatrix}
    \Re {\tilde \Sigma}_z & -\Im {\tilde \Sigma}_z\\
    \Im {\tilde \Sigma}_z & \Re {\tilde \Sigma}_z\\
  \end{bmatrix}, \quad z = 1, \ldots, K,
\]
the real-valued transformation of $\tilde \Sigma_z$ \cite[Ch. 4]{boyd2004convex}.
(Our model estimates the collection of real-valued
natural parameters $\theta = (\Sigma^{-1}_1, \ldots, \Sigma^{-1}_K)$;
it is trivial to obtain the equivalent collection of complex-valued natural parameters.)
For the remainder of this section, we only consider the problem in its real-valued form.

We generate approximately 2900 samples and randomly partition the data set into 
80\% training samples and 20\% test samples.
The number of training samples per vertex vary significantly;
there are a mean of 1.74 samples per vertex,
and the maximum number of samples on a vertex is 30.
625 of the $K=1000$ vertices have no training samples associated with them.

\paragraph{Model.}
The stratified model in this case is $K=1000$ (inverse) covariance matrices
in $\symm_{++}^{30}$, indexed by the radar orientation.
Our model has $Kn(n-1)/2=435000$ parameters.

\paragraph{Regularization.}
For local regularization, we utilize trace regularization with regularization 
weight $\gamma_{\mathrm{tr}}$, 
and $\ell_1$-regularization on the off-diagonal elements with regularization weight
$\gamma_{\mathrm{od}}$. 
That is, 
$r(\theta) = \gamma_{\mathrm{tr}}\Tr(\theta) + \gamma_{\mathrm{od}} \|\theta\|_{\mathrm{od},1}$.

The regularization graph for the stratified model is taken as the Cartesian product 
of three regularization graphs:
\begin{itemize}
	\item \emph{Range.} The regularization graph is a path graph with 10 vertices, 
	with edge weight $\gamma_\mathrm{range}$.
	\item \emph{Azimuth.} The regularization graph is a cycle graph with 10 vertices, 
	with edge weight $\gamma_\mathrm{azi}$.
	\item \emph{Doppler frequency.} The regularization graph is a path graph with 10 vertices, 
	with edge weight $\gamma_\mathrm{dopp}$.
\end{itemize}
The corresponding Laplacian matrix has 6600 nonzero entries and the hyper-parameters are 
$\gamma_\mathrm{range}$, $\gamma_\mathrm{azi}$, and $\gamma_\mathrm{dopp}$.

The stratified model in this case has five hyper-parameters: two for the local regularization,
and three for the Laplacian regularization graph edge weights.

\paragraph{Results.}
We compared a stratified model to a common model.
The common model corresponds to solving one individual covariance estimation problem, 
ignoring the radar orientations.
For the common model, we let $\gamma_{\mathrm{tr}}=0.001$ and $\gamma_{\mathrm{od}}=59.60$.
For the stratified model, we let $\gamma_{\mathrm{tr}}=2.68$, $\gamma_{\mathrm{od}}=0.66$, 
$\gamma_\mathrm{range} = 10.52$, $\gamma_\mathrm{azi} = 34.30$, and $\gamma_\mathrm{dopp} = 86.97$.
These hyper-parameters were chosen by performing a crude hyper-parameter search
and selecting hyper-parameters that performed well on the validation set.
We compare the models' average loss over the held-out test sets in table~\ref{tab:radar}.
In addition, we also compute the metric 
\[
D(\theta) = \frac{1}{Kn} \sum_{k=1}^K\left(\Tr(\Sigma_k^{\star}\theta_k) - \log\det(\theta_k) \right),
\]
where $\Sigma_k^{\star}$ is the true covariance matrix for the stratification 
feature value $z=k$;
this metric is used in the radar signal processing literature 
as a metric to determine how close $\theta_k^{-1}$ is to $\Sigma_k^{\star}$.

\begin{table}
\caption{Results for \S\ref{ss:radar}.}
 \vspace{.4em}
 \centering
 \begin{tabular}{lll}
   \toprule
   Model & Average test sample loss & $D(\theta)$\\
   \midrule
   Common & 0.153 & 2.02\\
   Stratified & 0.069 & 1.62\\
   \bottomrule
 \end{tabular}
 \label{tab:radar}
\end{table}

\paragraph{Application.}
As another experiment, we consider utilizing these models in a target 
detection problem: given a vector of data $y \in \reals^{30}$ and its 
radar orientation $z$, determine if the vector is just interference,
\ie,
\[
  y \mid z = d, \quad d \sim {\mathcal N}(0, \Sigma_z^{\star}),
\] 
or if the vector has some target associated with it, \ie,
\[
  y \mid z = s_z + d, \quad d \sim {\mathcal N}(0, \Sigma_z^{\star})
\]
for some target vector $s_z \in \reals^{30}$, which is fixed for each $z$.
(Typically, this is cast as a hypothesis test where the former is the null hypothesis
and the latter is the alternative hypothesis~\cite{ward1995stap}.)
We generate $s_z$ with $z=(r,a,d)$ as
\[
  s_z = (\Re \tilde{s}_z, \Im \tilde{s}_z ),
  \qquad \tilde{s}_z = (1, z_d, z_d^2) \otimes (1, z_a, z_a^2, z_a^3, z_a^4)
\]
with $z_a = e^{2 \pi i \sin(a)}$, $z_d = e^{2 \pi i d / f_R}$, and $f_R=1984$ is the 
pulse repetition frequency (in Hz);
these values are realistic and selected from the radar signal processing 
literature~\cite[Ch. 2]{kang2015stap}.
For each $z$, we generate $y$ as follows: 
we sample a $d \sim {\mathcal N}(0, \Sigma_z^{\star})$,
and with probability 1/2 we set $y = s_z + d$, and set $y = d$ otherwise. 
(There are 1000 samples).
We then test if $y$ contains the target vector via the selection criterion 
\[
  \frac{(s_z^T \theta_z y)^2}{s_z^T \theta_z s_z} > \alpha,
\]
for some threshold $\alpha$; this is well-known in the radar signal processing literature
as the optimal method for detection in this setting~\cite{robey1992stap, wicks2006stap, kang2015stap}. 
If the selection criterion holds, then we classify $y$ as containing a target;
otherwise, we classify $y$ as containing noise.

\begin{figure}
  \centering
   \includegraphics[width=.5\textwidth]{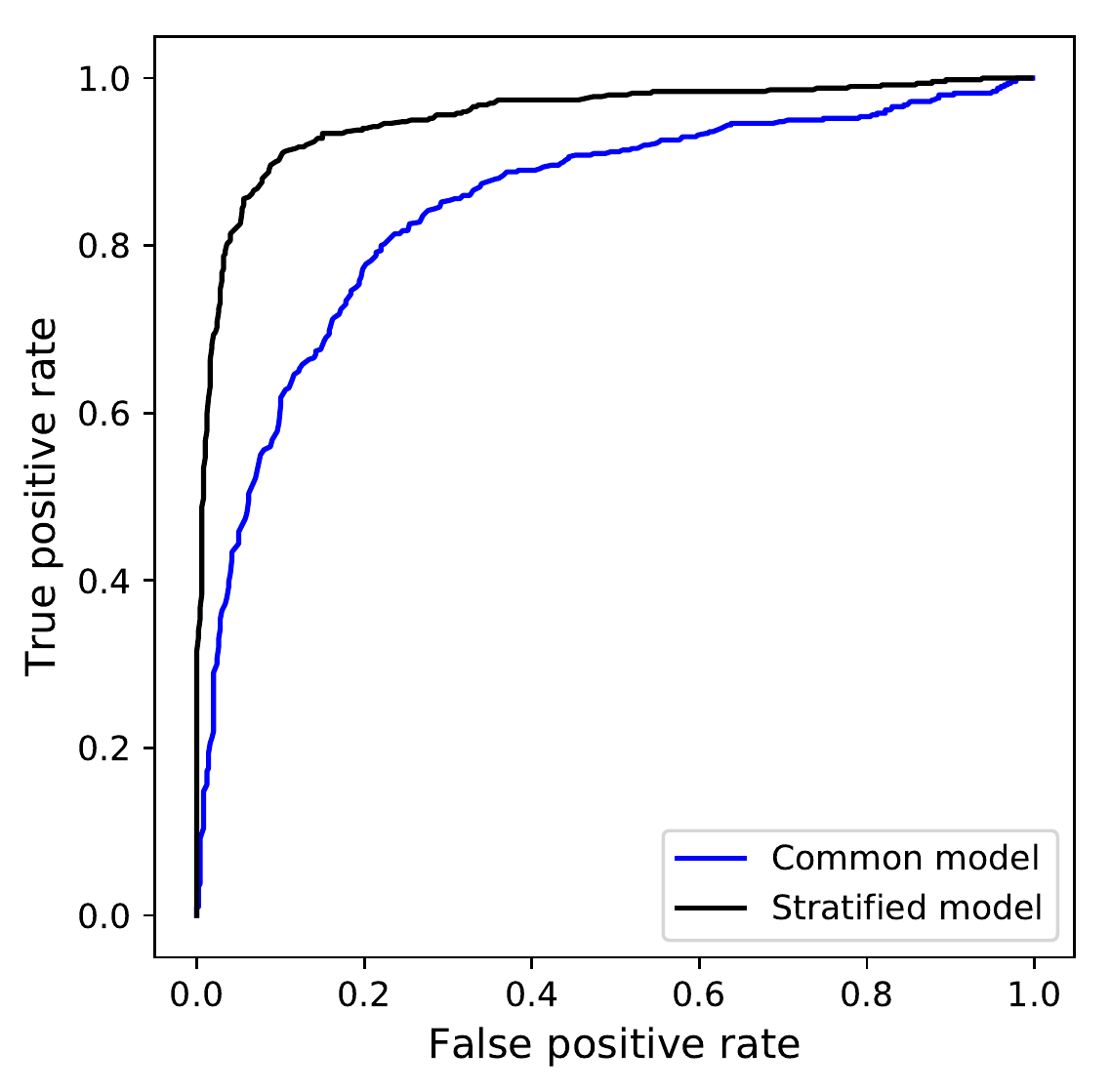}
     \caption{ROC curves for the common and stratified models as the threshold 
     $\alpha$ varies.}
  \label{fig:ROC_radar}
  \end{figure}

We vary $\alpha$ and test the samples on the common and stratified models.
We plot the receiver operator characteristic (ROC) curves for both models 
in figure~\ref{fig:ROC_radar}.
The area under the ROC curve is 0.84 for the common model and 0.95 for the 
stratified model; the stratified model is significantly more capable 
at classifying in this setting.

\clearpage
\subsection{Temperature covariance estimation}\label{ss:temperature}
We consider the problem of modeling the covariance matrix of hourly temperatures
of a region as a function of day of year.

\paragraph{Data records and dataset.}
We use temperature measurements (in Fahrenheit) 
from Boston, MA, sampled once per hour from October 2012 to October 2017, 
for a total of 44424 hourly measurements.
We winsorize the data at its 1st and 99th percentiles.
We then remove a baseline temperature, which consists of 
a constant and a sinusoid with period one year.
We refer to this time series as the baseline-adjusted temperature.

From this data, we create data records $(z_i,y_i), i = 1, \ldots, 1851$
(so $m=1851$),
where $y_i \in \reals^{24}$ is the baseline-adjusted temperature
for day $i$, and $z_i \in \{1, \ldots, 366\}$ is the day of the year.
For example, $(y_i)_3$ is the baseline-adjusted temperature at 3AM, 
and $z_i=72$ means that the day was the 72nd day of the year.
The number of stratification features is then $K=366$, corresponding
to the number of days in a year.

We randomly partition the dataset into a training set consisting of 60\% 
of the data records, a validation set consisting of 20\% of the data records, 
and a held-out test set consisting of the remaining 20\% of the data records.
In the training set, there are a mean of approximately 3.03 data records per 
day of year, the most populated vertex is associated with six data records, and
there are seven vertices associated with zero data records.

\paragraph{Model.}
The stratified model in this case is $K=366$ (inverse) covariance matrices
in $\symm_{++}^{24}$, indexed by the days of the year.
Our model has $Kn(n-1)/2=101016$ parameters.

\paragraph{Regularization.}
For local regularization, we utilize trace regularization with regularization 
weight $\gamma_{\mathrm{tr}}$, 
and $\ell_1$-regularization on the off-diagonal elements with regularization 
weight $\gamma_{\mathrm{od}}$. 
That is, 
$r(\theta) 
= \gamma_{\mathrm{tr}}\Tr(\theta) 
+ \gamma_{\mathrm{od}} \|\theta\|_{\mathrm{od},1}$.

The stratification feature stratifies on day of year;
our overall regularization graph, therefore, is a
cycle graph with 366 vertices, one for each possible day of the year,
with edge weights $\gamma_{\mathrm{day}}$.
The associated Laplacian matrix contains 1096 nonzeros.

\paragraph{Results.}
We compared a stratified model to a common model.
The common model corresponds to solving one covariance estimation problem, 
ignoring the days of the year.

For the common model, we used 
$\gamma_{\mathrm{tr}}=359$ and $\gamma_{\mathrm{od}}=0.1$.
For the stratified model, we used $\gamma_{\mathrm{tr}}=6000$, 
$\gamma_{\mathrm{od}}=0.1$, and $\gamma_{\mathrm{day}}=0.14$.
These hyper-parameters were chosen by performing a crude hyper-parameter search
and selecting hyper-parameters that performed well on the validation set.

\begin{figure}
\centering
 \includegraphics[width=.85\textwidth]{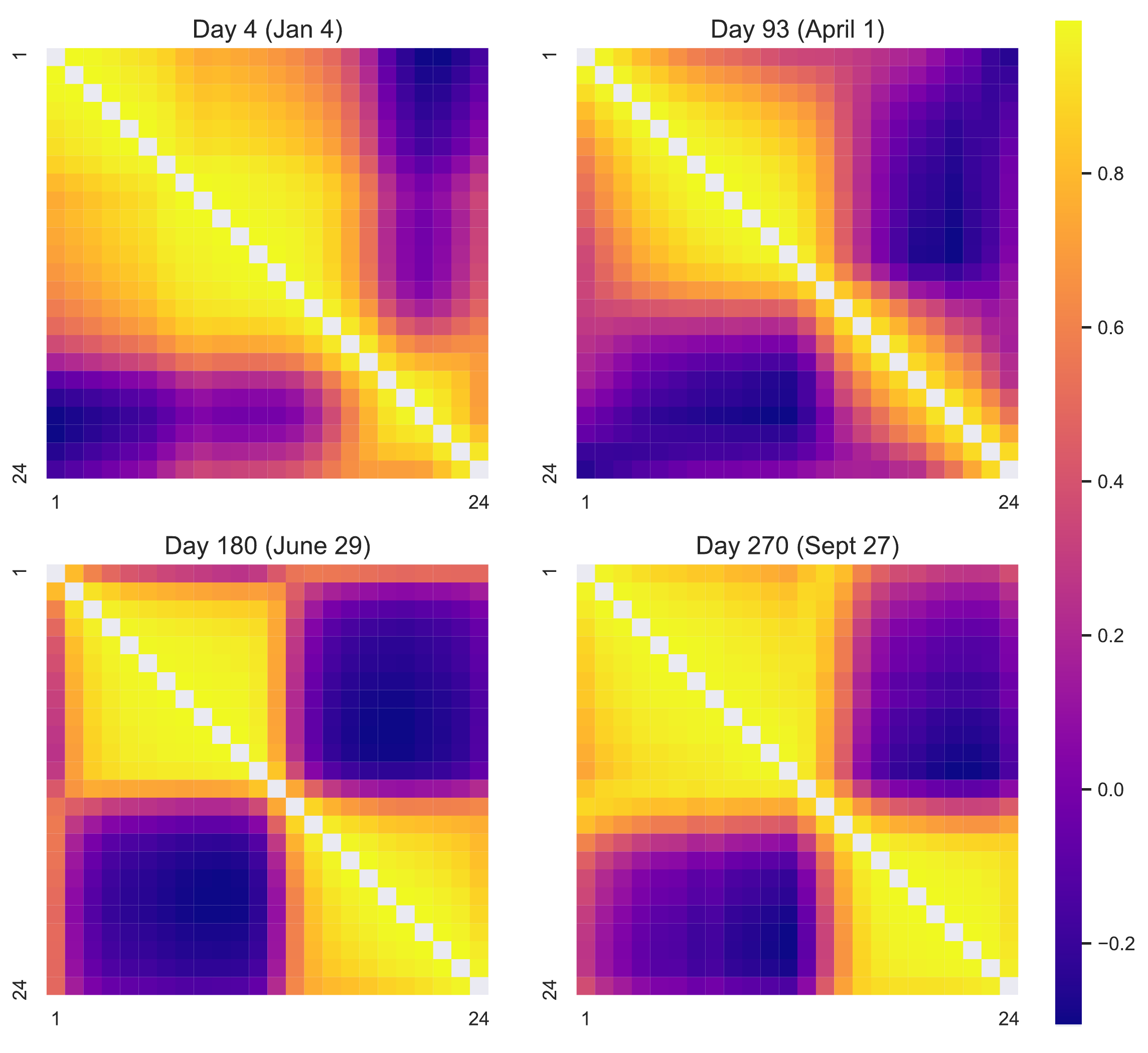}
   \caption{Heatmaps of the correlation matrices for days approximately corresponding to 
   the start of winter (top left), spring (top right), summer (bottom left) and autumn 
   (bottom right).}
\label{fig:corr_heatmaps_weather}
\end{figure}

We compare the models' losses over the held-out test sets in table~\ref{tab:temperature_exp1}.
\begin{table}
  \caption{Average loss over the test set for \S\ref{ss:temperature}.}
  \vspace{.4em}
 \centering
 \begin{tabular}{ll}
   \toprule
   Model & Average test loss \\
   \midrule
   Common & 0.132\\
   Stratified & 0.093\\
   \bottomrule
 \end{tabular}
 \label{tab:temperature_exp1}
\end{table}
To illustrate some of these model parameters, in figure~\ref{fig:corr_heatmaps_weather}
we plot the heatmaps of the correlation matrices for days that roughly correspond to 
each season.

\begin{figure}
  \centering
   \includegraphics[width=0.9\textwidth]{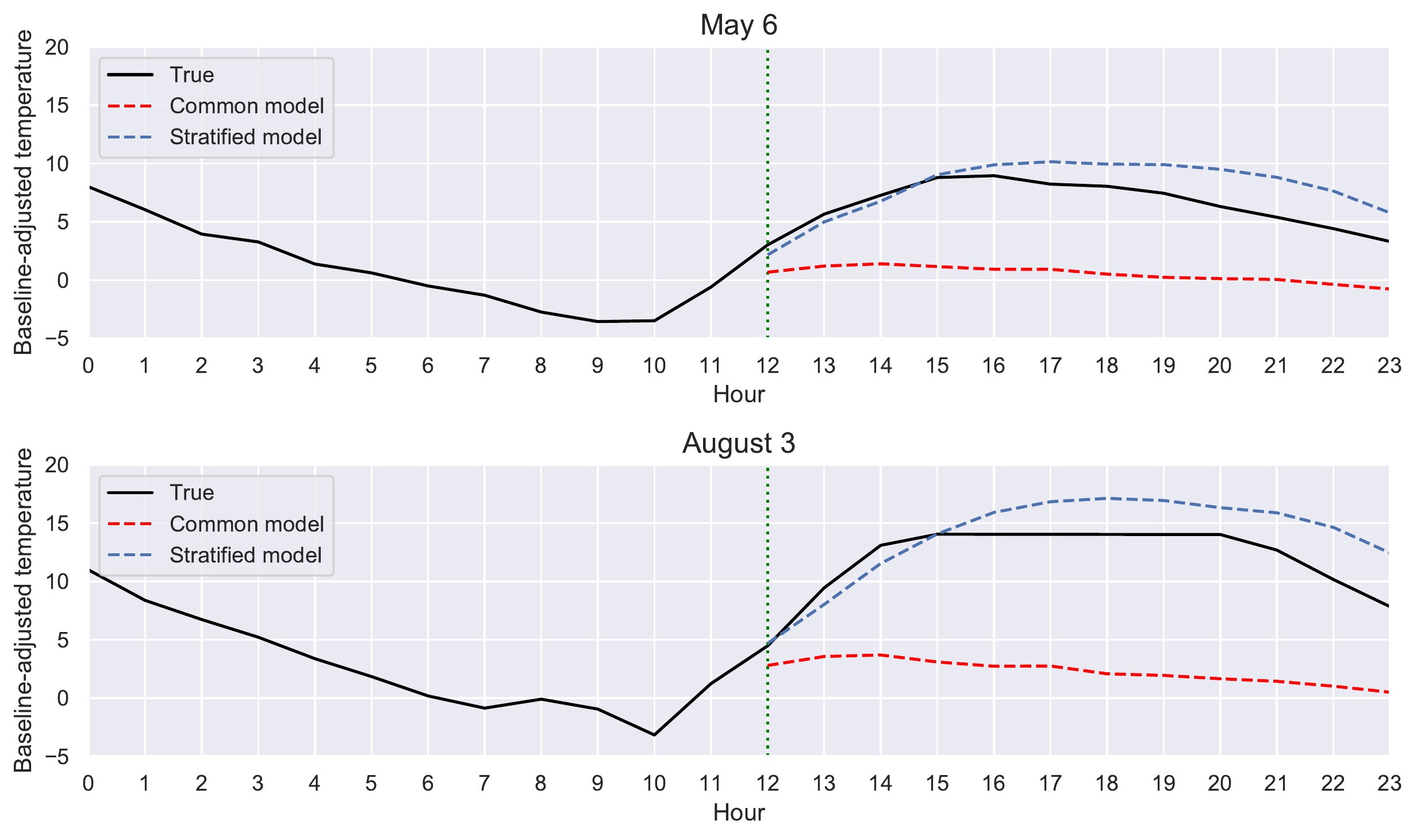}
     \caption{Baseline-adjusted temperature forecasts for two days 
     in the held-out test set.}
  \label{fig:temperature_preds_weather}
\end{figure}

\paragraph{Application.}
As another experiment, we consider the problem of forecasting the second half
of a day's baseline-adjusted temperature given the first half of the day's 
baseline-adjusted temperature.
We do this by modeling the baseline-adjusted temperature 
from the second half of the day as a Gaussian distribution conditioned on the observed 
baseline-adjusted temperatures~\cite{marriott1984gaussian,bishop2006gaussian}.
We run this experiment using the common and stratified models found in 
the previous experiment, using the data in the held-out test set.
In table~\ref{tab:temperature_exp2}, we compare the root-mean-square error 
(RMSE) between the predicted temperatures and the true temperatures over the 
held-out test set for the two models,
and in figure~\ref{fig:temperature_preds_weather}, we plot the temperature 
forecasts for two days in the held-out test set.
\begin{table}
\caption{Average prediction RMSE over the test set for \S\ref{ss:temperature}.}
 \vspace{.4em}
 \centering
 \begin{tabular}{ll}
   \toprule
   Model & {Average prediction RMSE} \\
   \midrule
   Common & 8.269\\
   Stratified & 6.091\\
   \bottomrule
 \end{tabular}
 \label{tab:temperature_exp2}
\end{table}

\section*{Acknowledgments} 
Jonathan Tuck is supported by the Stanford Graduate Fellowship in Science and Engineering.
The authors thank Muralidhar Rangaswamy and Peter Stoica for helpful comments on an 
early draft of this paper.


\newpage
\bibliography{cov_strat_models}
\end{document}